\pdfoutput=1

\documentclass[11pt]{article}

\usepackage[]{ACL2023}

\usepackage{booktabs}
\usepackage{graphicx}
\usepackage[normalem]{ulem}
\useunder{\uline}{\ul}{}
\usepackage{lscape}
\usepackage{booktabs}
\usepackage[normalem]{ulem}
\useunder{\uline}{\ul}{}
\usepackage{longtable}

\usepackage{times}
\usepackage{latexsym}
\usepackage{booktabs}
\usepackage{graphicx}
\usepackage{lipsum}
\usepackage[T1]{fontenc}
\usepackage{multirow}

\usepackage[utf8]{inputenc}
\usepackage{cleveref}

\usepackage{microtype}
\usepackage{comment}

\usepackage{inconsolata}

\title{SummQA at MEDIQA-Chat 2023: \\
In-Context Learning with GPT-4 for Medical Summarization}

\author{Yash Mathur\thanks{\enspace Equal contribution} \ \ \ Sanketh Rangreji\footnotemark[1]  \ \ \ Raghav Kapoor\footnotemark[1] \\ \textbf{Medha Palavalli\footnotemark[1] \ \ \ Amanda Bertsch \ \ \ Matthew R. Gormley}\\ 
 Carnegie Mellon University \\
{\tt\small [ymathur, srangrej, raghavka, mpalaval, abertsch, mgormley] @andrew.cmu.edu }}

\begin{document}
\maketitle

\begin{abstract}
Medical dialogue summarization is challenging due to the unstructured nature of medical conversations, the use of medical terminology in gold summaries, and the need to identify key information across multiple symptom sets. 
We present a novel system for the Dialogue2Note Medical Summarization tasks in the MEDIQA 2023 Shared Task. Our approach for section-wise summarization (Task A) is a two-stage process of selecting semantically similar dialogues and using the top-$k$ similar dialogues as in-context examples for GPT-4. For full-note summarization (Task B), we use a similar solution with $k$=1. We achieved 3rd place in Task A (2nd among all teams), 4th place in Task B Division Wise Summarization (2nd among all teams), 15th place in Task A Section Header Classification (9th among all teams), and 8th place among all teams in Task B. Our results highlight the effectiveness of few-shot prompting for this task, though we also identify several weaknesses of prompting-based approaches. We compare GPT-4 performance with several finetuned baselines. We find that GPT-4 summaries are more abstractive and shorter. We make our code publicly available \footnote{\url{https://github.com/Raghav1606/SummQA}}.
\end{abstract}

\section{Introduction}
Medical dialogue summarization is a long-standing challenge in NLP \cite{lopez-espejel-2019-automatic, joshi2020dr, chintagunta-etal-2021-medically, navarro-etal-2022-shot}. Medical scribes write notes on doctor-patient conversations in a predefined template called SOAP notes \cite{ullman2021scribes, podder2021soap}, which contains sections for information from the patient, test results and observations, diagnosis, and the conclusion or treatment. 

Medical summarization is challenging for several reasons. It requires dialogue understanding, where data is often limited \cite{dai-etal-2020-learning, lin2020graphevolving}; this is compounded by the sensitive nature of medical information, which restricts the release of training data for this task \cite{johnson2023mimic}. Doctors and patients may discuss several conditions in the same conversation, requiring the scribe to differentiate \cite{gidwani2017scribes, mishra2018scribes}. 
Scribes often use medical terminology in the notes that are not present in the doctor-patient conversation \cite{corby2020scribe}.
Additionally, medical summarization is a high-stakes domain \cite{naik2022ethics}, motivating several efforts to build explainabile systems for this task \cite{jain2022survey, reddy2022explain}.
In parallel, research on large language models (LLMs) has demonstrated compelling few-shot capabilities across domains \cite{brown2020fewshot, perez2021true}.

In this paper, we explore several potential applications of a recent LLM, GPT-4 \cite{openai2023gpt4}, on medical summarization. We use GPT-4 and finetuned BioBERT \cite{lee2020biobert} as an ensemble for classifying the section headers of medical summaries, a 20-category classification problem.
Then, given a candidate section header, we apply Maximal Marginal Relevance (MMR) \cite{carbonell1998use} to select examples for a fewshot demonstration and use these examples to prompt GPT-4 for section-wise summarization. 
This approach outperforms finetuning BART \cite{lewis2019bart} and T5 \cite{raffel2020exploring} over the limited available data.
For full-note summarization, we take a similar approach, but select only a single example for the demonstration due to the increased length of the inputs.
This also outperforms our supervised baselines.
We outline several additional potential prompting approaches and compare their relative efficacy. 

\begin{figure*}
  \includegraphics[scale=0.5]{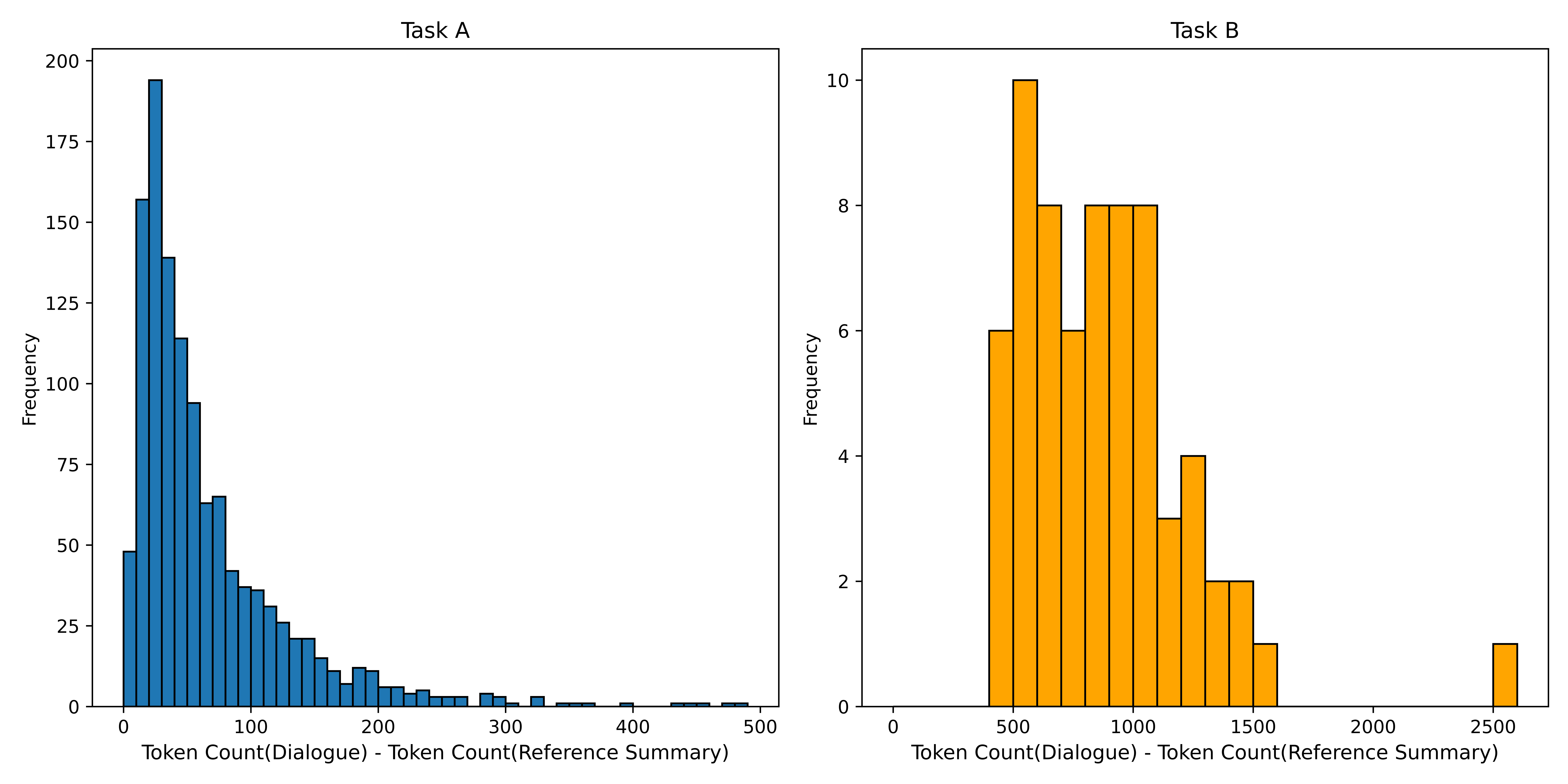}
  \caption{Distribution of difference in length between dialogue and reference summary. A larger difference in length indicates a higher degree of compression.}
  \label{fig:dataset_diff}
\end{figure*}

Applying LLMs for medical summarization is a compelling solution to the data scarcity problems in this domain, and we find promising performance, with our team placing second in the MEDIQA 2023 Shared Task for Subtask A and Division Summary for Subtask B. 
However, we also identify key areas for improvement. 
We analyze the differences in outputs between the settings in output length and extractive ability. We find that the summaries generated by LLMs tend to be shorter and less extractive than human-generated summaries as well as SOTA fine-tuned biomedical summarization models.
We also note the impracticality of this approach for real data, due to privacy concerns.

\begin{figure}
  \includegraphics[scale=0.45]{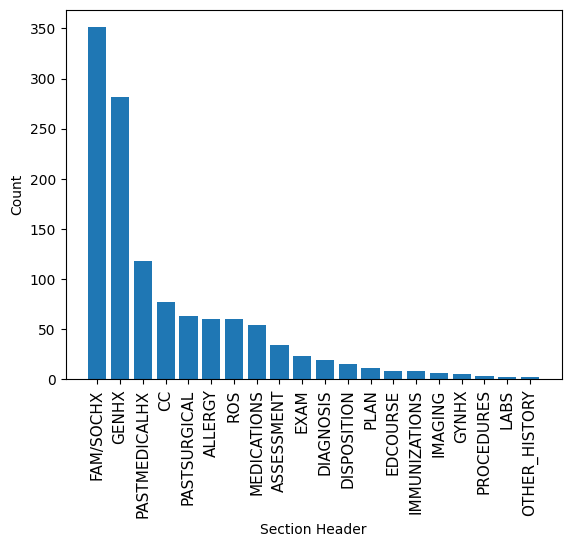}
  \caption{Section header distribution for Task A.}
   \label{fig:section_dist}
\end{figure}

\section{Background}
Dialogue2Note Summarization was one of two tracks in the  MEDIQA-Chat 2023 shared task \cite{mediqa-chat-2023}. 
The track was further comprised of two tasks. 

\paragraph{Task A} involves generating a section-specific clinical summary from a conversation between a patient and a doctor. Additionally, Task A includes a classification task: assigning each dialogue an appropriate section header. There are 1200 conversations in the training split of the dataset \cite{mts-dialog}  for Task A. The distribution over the section headers in Task A is a long-tailed distribution, displayed in Figure \ref{fig:section_dist}.

\paragraph{Task B} involves generating a full note summary given a conversation; these summaries were evaluated on the section-level and the full-note level. There were 67 conversations in the training split of the dataset \cite{aci-demo}; these dialogues and reference summaries are significantly longer than those for Task A, as these dialogues encompass an entire conversation between a patient and a doctor. The distribution of the difference in dialogue and summary length for both tasks is shown in Figure \ref{fig:dataset_diff}.


\begin{figure*}
  \includegraphics[width=\textwidth]{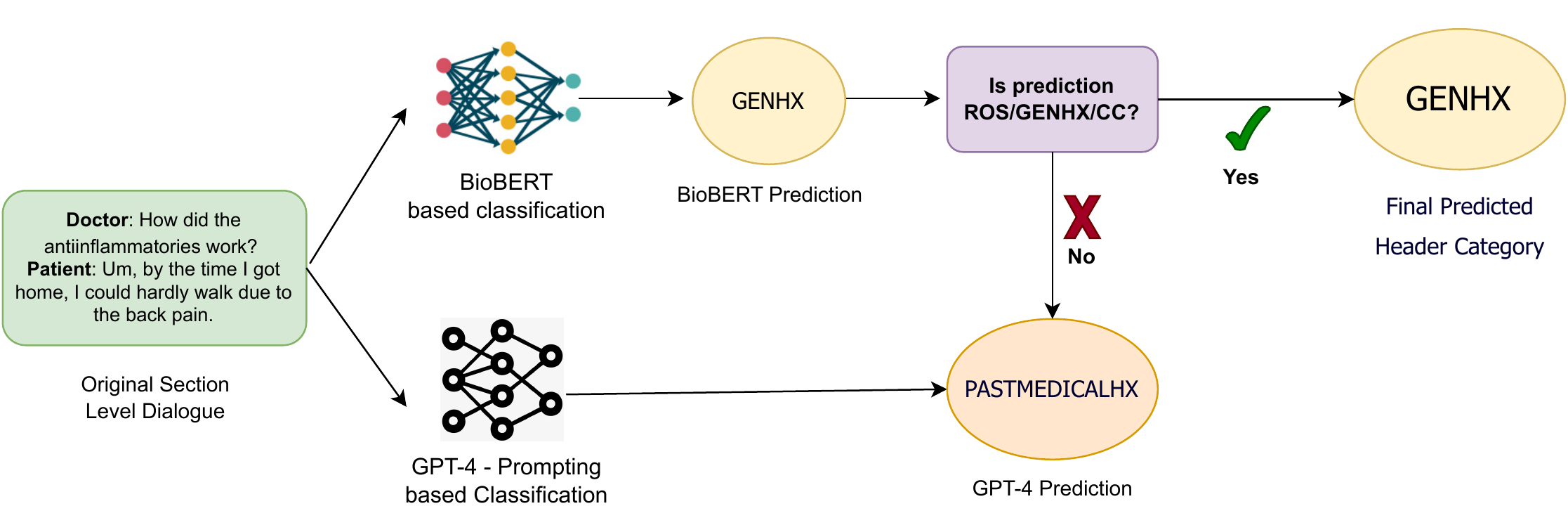}
  \caption{System Architecture for Section Header Classification (Task A)}
  \label{fig:class_arch}
\end{figure*}

\section{Related Work}

\paragraph{Summarization} 
In recent years, fine-tuning pre-trained models on domain-specific datasets has been the leading practice in text summarization research. While these models produce high-quality summaries and earn high scores against standard benchmarks, they require large datasets in order to adapt to specific domains or summarization styles \cite{lewis2020BART}. Transformer-based models \cite{michalopoulos-etal-2022-medicalsum} and pointer generator network models \cite{joshi-etal-2020-dr} have been fine-tuned with medical domain knowledge to produce summaries that achieve state-of-the-art results. 

Maximal Marginal Relevance was created to reduce redundancies in multi-document summaries \cite{goldstein-carbonell-1998-summarization}. \citet{abdullah2023MMR} used MMR to generate query-focused summaries from pre-trained models without performing fine-tuning. \cite{ye2022complementary} use MMR to select examples for in-context prompting. 

The success of prompt-based models such as GPT-3 \cite{brown2020fewshot} has allowed for learning from natural language task instructions and/or a few demonstrative examples in the context without updating model parameters. 
In news summarization, \citet{goyal2022news} find that GPT-3 summaries were preferred by humans over summaries from fine-tuned models trained on large summarization datasets; they posit that zero-shot summaries avoid pitfalls from low-quality training data that are common in summaries from fine-tuned models.
In the biomedical domain, pre-trained language models and few-shot learning has been used to collect and generate labeled data for medical dialogue summarization \cite{chintagunta-etal-2021-medically}. 
Recent work has used GPT-4 to pass the USMLE without any specialized prompt crafting \cite{nori2023capabilities} and perform zero-shot medical evidence summarization across six clinical domains \cite{tang-etal-2023-evaluating}.

\paragraph{Few-shot learning}

\begin{figure*}
  \includegraphics[width=\textwidth]{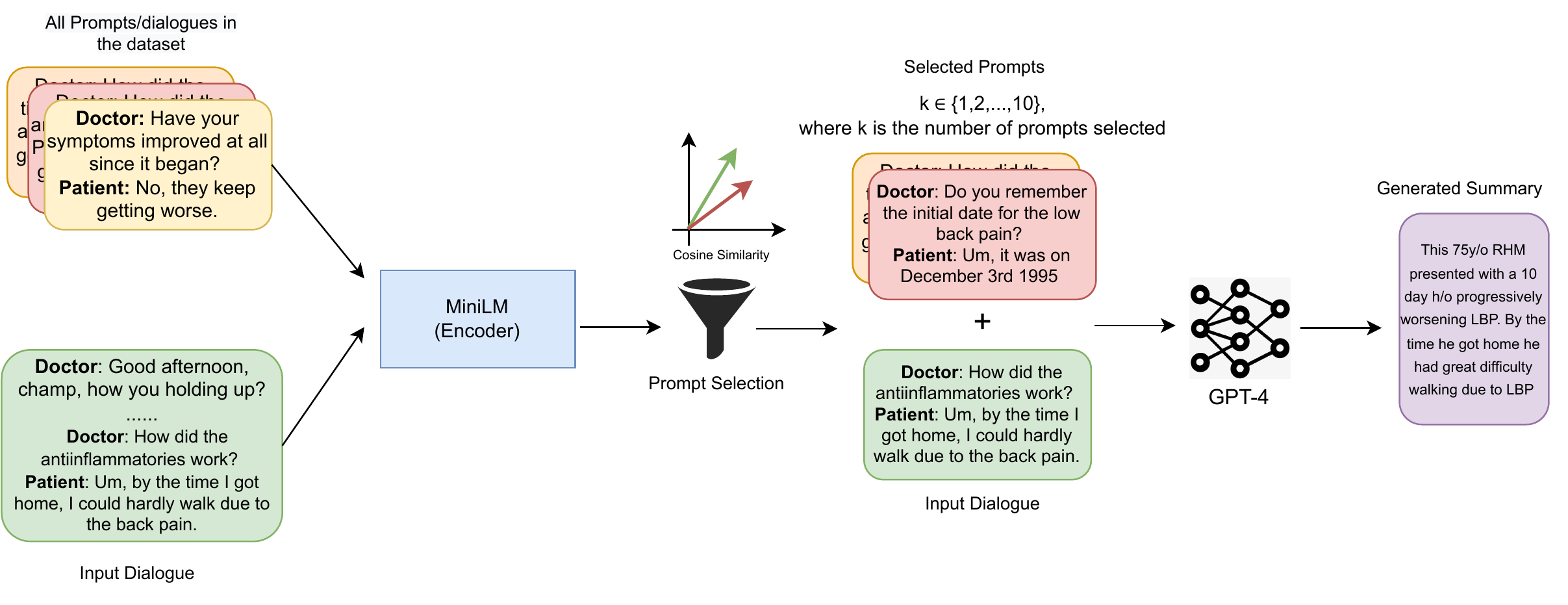}
  \caption{System Architecture for Summarization Task}
  \label{fig:sys_arch_summ}
\end{figure*}


Few-shot learning can be unstable as the prompt format, training examples, and even the order of the training examples can cause accuracy to vary from near chance to near state-of-the-art \cite{lu-etal-2022-fantastically}. Recent work on prompting has tried to mitigate these problems through techniques such as calibration \cite{zhao2021calibrate}, prompt combination \cite{zhou2022promptconsistency}, or automatic prompt generation \cite{gao2021learners}. 

To mitigate any instability caused by a model's bias, \citet{zhao2021calibrate} estimated the bias towards each answer by asking for its prediction when given the training prompt and a content-free test input such as “N/A” and then fit calibration parameters that cause the prediction for this input to be uniform across answers. To date, studies in prompt combination are rooted in paraphrasing-based methods that take a seed prompt and paraphrase it into several semantically similar expressions. Typically simple ensemble methods \cite{zhou2022promptconsistency} such as Maximal Marginal Relevance \cite{mao-etal-2020-multi} are used to combine the answers to the different prompts as to provide each prompt to contribute to the final answer. 

A number of techniques have also been proposed for selecting fewshot examples \cite{rubin-etal-2022-learning}. Fewshot techniques often rely on selecting optimal examples from a large dataset; some work has shown that this leads to an overstatement of fewshot performance, as a large number of labeled examples are necessary to select good examples for the fewshot prompt \cite{perez2021true}. We note that we use the full datasets (1,200 examples for Task A, 40 for Task B) for our prompt selection techniques.

\section{Methodology and Baselines}
Our summary generation pipeline remains the same across the two tasks: we use GPT-4 to generate a summary given $k$ in-context examples.

\subsection{Task A - Section Level Summary}

Task A is composed of two subtasks, namely the section header classification and the section-level summarization. We discuss our approach for each of the subtasks below.

\subsubsection{Section Header Classification}
For the section header classification task, we create an ensemble of two models: BioBERT  \cite{lee2020biobert} and GPT-4. We fine-tune BioBERT with the training data provided for task A. We leverage GPT-4 to perform zero-shot classification on a sample with a given prompt (shown in Table \ref{tab:prompts}). During our analysis of each model's performance, we observe that each model is more accurate than the other on a distinct subset of classes. To leverage the varying nature of predictions from the models we build an ensemble classifier. The overall accuracies are shown in Table \ref{tab:my-table}. We observe empirically that our prompting-based approaches do not perform well on three of the section headers: ROS (Review of Systems), GENHX (History of the present illness), and CC (Chief Complaint), To create an ensemble classifier, we select BioBERT's predictions when it classifies a dialogue as one of these three section headers, and we select the prediction of our GPT-4 based pipeline otherwise. We present the architecture of our final model in Figure \ref{fig:class_arch}.

\subsubsection{Section Summary} 
\label{taskA_section_summary}
To generate summaries for a given section, we follow a multi-step process as shown in Figure \ref{fig:sys_arch_summ}. We encode each dialogue in the training data with MiniLM \cite{wang2020minilm}. For each dialogue to be summarized, we calculate cosine similarities with encoded dialogues from the training data. We retrieve $k$=7 similar examples from the training data based on the highest similarity. This similarity search, using cosine-similarity, serves as a prompt selection method, and the resulting few-shot prompts, with $k$=7 are then fed to GPT-4 along with their section headers to obtain a summary for the given section. We provide the prompt templates used in Table \ref{tab:prompts}. We select $k$=7 as this fits well in the context length of our prompting-based pipeline; we perform an analysis with varying $k$ in section \ref{sec:varying-k}.

\subsection{In-context Example Selection for Summarization Tasks}
This approach involves the dynamic selection of in-context examples for each example during validation or testing. This process entails utilizing matching or similarity criteria to compare the input dialogue of a specific example to a candidate pool comprising the complete training set. Through this process, we are able to select the most suitable examples for each individual case, thereby enhancing the efficacy of our prompts.

\paragraph{Semantic Similarity}
Here we select the $k$ most similar examples (dialogue and summary pairs) based on semantic similarity between the provided input dialogue and the input dialogues in the training set. We store the selected examples and generate prompts which we then evaluate on the validation/test set.

\paragraph{Maximal Marginal Relevance}
We select $k$ few-shot prompts using Maximal Marginal Relevance (MMR). Similarly to \citet{ye2022complementary}, we use MMR to select an example and use it as a one-shot example for prompting. 
Our choice of MMR was motivated by the idea that the diversity in the selected in-context examples of the prompt would help with some generalization;.

\subsection{Task B - Full Note Summary}
For the summarization of entire dialogues, our goal is to generate a full note containing all the appropriate sections. We use a similar approach as described in section \ref{taskA_section_summary} but we restrict it to $k$=1 similar examples from the training set and include section-level headers in the prompts to help the model understand the sections in the sumary. We selected one in-context example due to long example length relative to the context window of the model. The one-shot prompt is then again fed to the GPT-4 model to obtain a full-length summary. The results f


\begin{table*}[]
\centering
\resizebox{\textwidth}{!}{%
\begin{tabular}{ccccccccc}
\hline
\textbf{Task} &
  \textbf{Models} &
  \textbf{R1} &
  \textbf{R2} &
  \textbf{RL} &
  \textbf{BR} &
  \textbf{BP} &
  \textbf{BF1} &
  \textbf{BL} \\ \hline
 &
  Few-shot Text-davinci-003 &
  17.3 &
  6.8 &
  13.5 &
  0.562 &
  0.539 &
  0.546 &
  0.398 \\
   &
  T5-Small &
  30.0 &
  {\color[HTML]{212121} 11.3} &
  23.6 &
  0.631 &
  0.675 &
  0.646 &
  0.445 \\
A &
  Two Stage Prompting &
  28.4 &
  11.6 &
  21.6 &
  {\color[HTML]{212121} 0.694} &
  {\color[HTML]{212121} 0.633} &
  {\color[HTML]{212121} 0.656} &
  {\color[HTML]{212121} 0.547} \\
  &
  Prompt Selection text-davinci-003 (Semantic) &
  {\color[HTML]{212121} 38.6} &
  {\color[HTML]{212121} 18.4} &
  {\color[HTML]{212121} 31.2} &
  {\color[HTML]{212121} 0.716} &
  {\color[HTML]{212121} 0.725} &
  {\color[HTML]{212121} 0.715} &
  {\color[HTML]{212121} 0.56} \\
 &
  Prompt Selection text-davinci-003 (MMR) &
  39.2 &
  \textbf{18.8} &
  31.9 &
  0.717 &
  0.725 &
  0.716 &
  {\color[HTML]{212121} 0.559} \\
 &
  \textbf{Prompt Selection GPT4 (Semantic)} &
  {\color[HTML]{1D1C1D} \textbf{42.8}} &
  {\color[HTML]{1D1C1D} 17.2} &
  {\color[HTML]{1D1C1D} \textbf{32.3}} &
  {\color[HTML]{1D1C1D} \textbf{0.719}} &
  {\color[HTML]{1D1C1D} \textbf{0.729}} &
  {\color[HTML]{1D1C1D} \textbf{0.720}} &
  {\color[HTML]{1D1C1D} \textbf{0.564}} \\  \hline
  &
T5-Small &
  22.7 &
  {\color[HTML]{212121} 9.1} &
  12.9 &
  0.568 &
  0.471 &
  0.514 &
  0.319 \\
 B & Perspective Shift &
  {\color[HTML]{212121} 38.7} &
  {\color[HTML]{212121} 15.6} &
  {\color[HTML]{212121} 23.8} &
  {\color[HTML]{212121} 0.618} &
  {\color[HTML]{212121} 0.679} &
  {\color[HTML]{212121} 0.647} &
  {\color[HTML]{212121} 0.392} \\
  &
  Zero Shot Text-davinci-003 &
  {\color[HTML]{212121} 45.4} &
  {\color[HTML]{212121} 23.3} &
  {\color[HTML]{212121} 30.6} &
  {\color[HTML]{212121} 0.644} &
  {\color[HTML]{212121} 0.712} &
  {\color[HTML]{212121} 0.676} &
  {\color[HTML]{212121} 0.410} \\
 &
  \textbf{Prompt Selection GPT4 (k = 1)} &
  {\color[HTML]{1D1C1D} \textbf{50.7}} &
  {\color[HTML]{1D1C1D} \textbf{24.9}} &
  {\color[HTML]{1D1C1D} \textbf{33.6}} &
  {\color[HTML]{1D1C1D} \textbf{0.666}} &
  {\color[HTML]{1D1C1D} \textbf{0.703}} &
  {\color[HTML]{1D1C1D} \textbf{0.684}} &
  {\color[HTML]{1D1C1D} \textbf{0.406}} \\
\hline
\end{tabular}%
}
\caption{Validation Results for Task A and Task B Summarization. Metrics include ROUGE-1 (R1), ROUGE-2 (R2), ROUGE-L (RL), BERTScore Precision (BP), Recall (BR), and F1 (BF1), and BLEURT (BL).}
\label{tab:res_summ}
\end{table*}

\subsection{Baseline Approaches}

We also consider a variety of baseline approaches including, supervised fine-tuning of T5, zero-shot/few-shot GPT-3, perspective-shifting the dialogue followed by summarization, two-stage prompting, our similarity-based in-context learning applied to GPT-3, and mixing of extractive/abstractive methods.

\subsubsection{T5}

We fine-tuned the T5-small model for the end-end full-length summarization task (Task B). We finetuned for 20 epochs with a learning rate of 0.001. 
Our objective was to obtain a basic model that can serve as a benchmark to assess the complexity and difficulties associated with this specific task. 
We find that this finetuned model significantly underperforms our other methods, with a ROUGE-1 of 20.187; this may be due to the small dataset for finetuning or a non-optimized set of hyperparameters, as we do not do extensive hyperparameter search. 

\subsubsection{GPT-3}
We investigated several prompting strategies and approaches using \textit{text-davinci-003}.
\paragraph{Zero-shot prompting}
For Task B we used the prompt template mentioned in the Appendix \ref{sec:appendix}, where we specified the dialogue to be summarized with an instruction prompt mentioning the 4 main sections usually reported in the SOAP notes - "HISTORY OF PRESENT ILLNESS", "PHYSICAL EXAM", "RESULTS" and  "ASSESSMENT AND PLAN". The zero-shot prompt gave us a reasonably high ROUGE-1 score of 45.911.


\begin{table}[]
\centering
\begin{tabular}{lc}
\hline
\textbf{Model}                & \multicolumn{1}{l}{\textbf{Accuracy}} \\ \hline
GPT-3.5-turbo  & 68.943                                \\
GPT-4                         & 69.474                                \\
BioBERT                       & 71.278                                \\
Ensemble (GPT-4 + BioBERT)    & \textbf{75.312}  \\
\hline
\end{tabular}%
\caption{Validation Results for Header Classification}
\label{tab:my-table}
\end{table}

\subsubsection{Few-shot prompting for section-wise summary}
For Task A, we employed \texttt{text-davinci-003} few-shot prompting strategy. Initially, we grouped and categorized the existing 20 section headers for the dataset into 4 main sections, namely "HISTORY OF PRESENT ILLNESS", "PHYSICAL EXAM", "ASSESSMENT AND PLAN", and "RESULTS". The categorization scheme is detailed in Table \ref{tab:cat}. It is worth noting that "Medications" can be categorized under either "HISTORY OF PRESENT ILLNESS" or "ASSESSMENT AND PLAN" and therefore appears in both categories. We created four few-shot prompt templates, each comprising $k$=5 in-context examples, for each section. For each example in the validation set, we selected the appropriate prompt based on the classified section header. 

\paragraph{Perspective Shift}
In this method which we evaluated for Task B, we adopt a two-stage prompting approach where we first use gpt3.5-turbo to obtain a third-person narrative from the input dialogue, following \citet{bertsch-etal-2022-said}, and use the third-person perspective narrative generated as input to a \textit{text-davinci-003} model to generate a summary using the same instruction prompt specifying each section header that needs to be generated.

\paragraph{Two Stage Prompting}
In this approach we defined two chained prompts applied one after the other in a stage-wise manner. The first stage prompt was "List the important points from the above conversation for a medical report". This generated a list of salient points summarizing the dialogue. The second stage prompt we used was "Create a paragraph from the above facts only". The output from this prompt served as the final summary, which we then evaluated. We opted for these specific phrasings in the second prompt to mitigate the issue of model hallucination, which we observed was prevalent when tasked with generating a medical summary directly.


\section{Results and Analysis}

\subsection{Experimental Setup}
We used an 80/20 train/validation split on the training set and used the entire validation split as our test set. The main hyperparameter that we varied across our experiments for prompt selection was k, the number of in-context examples we selected for the prompt. We report the ablation study on varying k over the validation split in Table \ref{tab:ablation_k}. For generations we used a single decoding (n = 1), temperature = 1.0, $top_p$ = 1.0 and $max_{tokens}$ = 800. The metrics for BERTScore and BLEURT in Table \ref{tab:res_summ} have been calculated using RoBERTa Large  \cite{liu2019roberta} and BLEURT-Tiny\footnote{\url{https://github.com/google-research/bleurt}} respectively. 

\subsection{Experimental Results}
Our experiment involving prompt selection via semantic similarity with GPT-4 yielded the most favorable outcomes on the validation split, and prompt selection was the best approach for both Task A and Task B. We propose that the remarkable performance of prompt selection is attributed to the in-context examples that were selected using semantic similarity with the input dialogue. This approach facilitates the generation of an example-specific prompt that incorporates similar in-context examples, leading to an improvement in the model's ability to produce summaries that are more relevant and precise. The use of semantic similarity allows for the identification of examples that share similar semantic structures with the input dialogue, thereby increasing the likelihood of generating coherent and accurate summaries.

\subsection{Length of Generated Summary vs. Reference Summary} 
\begin{figure*}
\centering
  \includegraphics[width=\textwidth]{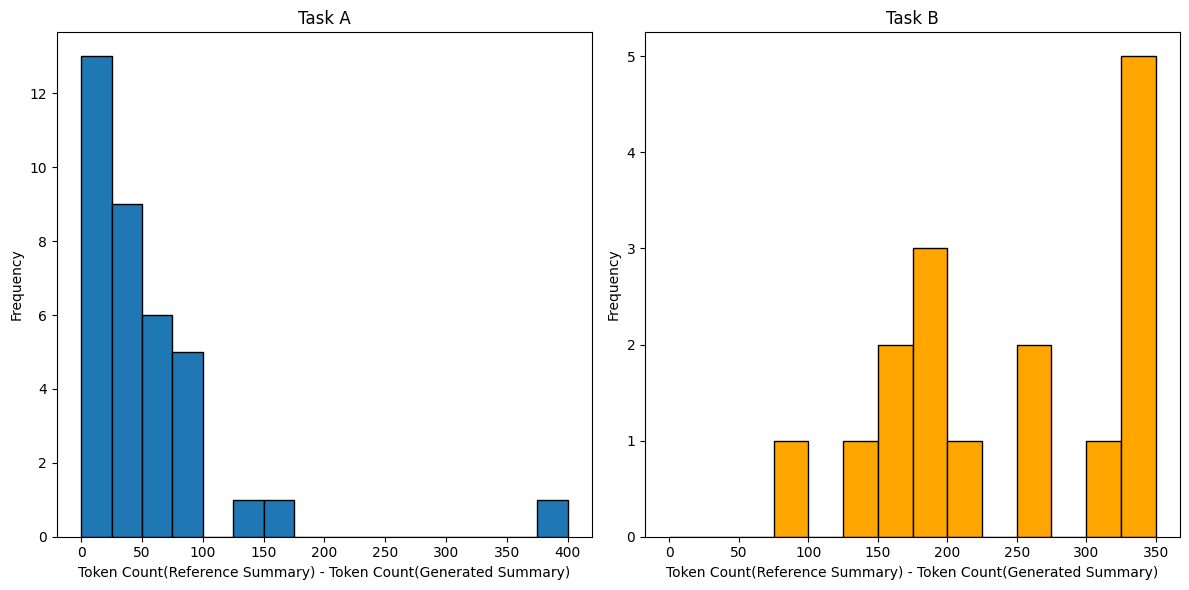}
  \caption{Difference in length of reference summaries and generated summaries}
  \label{fig:diff_ref}
\end{figure*}

As shown in Figure \ref{fig:diff_ref} we see that most generated summaries were shorter than reference summaries across tasks. This difference was more pronounced in Task B and therefore the summaries produced by our approach fall short in length thereby affecting the ROUGE-1 score as the number of matching n-grams is less. However, we observe that the BERT score still remains consistent even while producing shorter summaries. 

Another interesting observation is that individual section summaries, when combined together to produce a full-length summary are closer to the original length rather than prompting GPT-4 to generate a complete summary together. Hence, ensembling multiple section-level summaries to produce a longer summary is an approach we can explore further. We also tried multiple prompt templates (refer Table \ref{tab:prompts}), encouraging the model to produce longer summaries. However, the fact that we require a \textit{summary} induces the model to be concise.

\begin{table}[]
\centering
\begin{tabular}{ccccc}
\hline
\multicolumn{1}{l}{\textbf{Task}} & \textbf{Summary} & \textbf{EFC} & \textbf{EFD} & \textbf{CR} \\ \hline
\multirow{2}{*}{A} & Reference & 0.689 & 1.648 & 3.387 \\
                   & Generated & 0.561 & 1.036 & 5.701 \\ \hline
\multirow{2}{*}{B} & Reference & 0.671 & 2.044 & 2.856 \\
                   & Generated & 0.781 & 3.086 & 5.281 \\
\hline
\end{tabular}%
\caption{Summary extractiveness comparison - Extractive Fragment Coverage(EFC), Extractive Fragment Density(EFD), Compression Ratio (CR)}
\label{tab:extractive_task_A}
\end{table}

\subsection{Extractiveness of Summaries}
We measure the extractiveness of the generated summaries using three measures namely. extractive fragment coverage (EFC) \cite{grusky2020newsroom}, extractive fragment density (EFD) \cite{grusky2020newsroom} and compression ratio (CR) \cite{grusky2020newsroom}. These metrics rely upon the concept of extractive fragments which are defined as shared sequences in the dialogue and the summary. The extractive fragment coverage quantifies the percentage of words in the summary that are a part of the extractive fragments in the original dialogue. The extractive fragment density measures the average length of the extractive fragment to which a word in the summary belongs to. Compression is measured as the fraction of words in the article and summary. 

A comparison of the extractiveness and compression ratio of the reference and generated summaries are shown in Table \ref{tab:extractive_task_A}. For Task A, the relatively poor extractive performance of our proposed methods could be due to the smaller size of generated summaries which prevents the usage of more terms from the dialogue. In Task B, we observe the extractive capability of our model improving. This could be attributed to the larger dialogues in Task B allowing for a larger candidate space of tokens to be used in the generations. The generated summaries in Task B are still smaller than the reference summaries as shown by the compression ratio.

\begin{table*}[]
\centering
\begin{tabular}{clrrrrrrr}
\hline
Task &
  Runs &
  \multicolumn{1}{l}{\textbf{R1}} &
  \multicolumn{1}{l}{\textbf{R2}} &
  \multicolumn{1}{l}{\textbf{RL}} &
  \multicolumn{1}{l}{\textbf{BR}} &
  \multicolumn{1}{l}{\textbf{BP}} &
  \multicolumn{1}{l}{\textbf{BF1}} &
  \multicolumn{1}{l}{\textbf{BL}} \\ \hline
 &
  Run 1 &
  42.8 &
  17.2 &
  32.3 &
  0.719 &
  0.729 &
  0.720 &
  0.564 \\ 
 A &
  Run 2 &
  42.8 &
  17.1 &
  32.4 &
  0.718 &
  {\color[HTML]{212121} 0.727} &
  {\color[HTML]{212121} 0.717} &
  {\color[HTML]{212121} 0.558} \\  
&
  Run 3 &
  {\color[HTML]{212121} 42.6} &
  {\color[HTML]{212121} 17.1} &
  {\color[HTML]{212121} 32.1} &
  {\color[HTML]{212121} 0.716} &
  {\color[HTML]{212121} 0.724} &
  {\color[HTML]{212121} 0.714} &
  {\color[HTML]{212121} 0.551} \\ \hline
 &
  Run 1 &
  50.7 &
  {\color[HTML]{212121} 24.9} &
  33.6 &
  0.666 &
  0.703 &
  0.684 &
  0.406 \\ 
 B &
  Run 2 &
  {\color[HTML]{212121} 51.3} &
  {\color[HTML]{212121} 24.9} &
  {\color[HTML]{212121} 33.7} &
  {\color[HTML]{212121} 0.668} &
  {\color[HTML]{212121} 0.704} &
  {\color[HTML]{212121} 0.686} &
  {\color[HTML]{212121} 0.411} \\  
    &
  Run 3 &
  {\color[HTML]{212121} 50.9} &
  {\color[HTML]{212121} 24.8} &
  {\color[HTML]{212121} 33.6} &
  {\color[HTML]{212121} 0.666} &
  {\color[HTML]{212121} 0.704} &
  {\color[HTML]{212121} 0.684} &
  {\color[HTML]{212121} 0.407} \\
\hline
\end{tabular}%
\caption{Stability of Validation Results.  Metrics include ROUGE-1 (R1), ROUGE-2 (R2), ROUGE-L (RL), BERTScore Precision (BP), Recall (BR), and F1 (BF1), and BLUERT (BL).}
\label{tab:res_stability}
\end{table*}



\subsection{Impact of the number of in-context examples}
\label{sec:varying-k}
We further evaluate the impact of the number of in-context examples (k) on various metrics. We report the metrics in Table \ref{tab:ablation_k}. We observe a general improvement across all metrics as we increase k. This implies that the generated summaries improve as the model is given more access to relevant data through in-context examples. The relevancy in our method is brought in through the selection of in-context examples via semantic similarity/maximum marginal relevancy. This experiment was only performed for Task A since the token limits of the models did not permit the ablation of $k$ for Task B.

\begin{table}[]
\centering
\resizebox{7cm}{!}{%
\begin{tabular}{cccccc}
\hline
$k$&
  \textbf{R1} &
  \textbf{BP} &
  \textbf{BR} &
  \textbf{BF1} &
  \textbf{BL} \\ \hline
3 & 40.3 & 0.710 & 0.693 & 0.708 & 0.486 \\
5 & 41.9 & 0.718 & 0.706 & 0.714 & 0.521               \\
7 & 42.8 & 0.729 & 0.719 & 0.720 & 0.564       \\
\hline
\end{tabular}%
}
\caption{Impact of number of in-context examples ($k$) for Task A (GPT-4)}
\label{tab:ablation_k}
\end{table}

\section{Future Work and Conclusion}
This paper attempts to automatically generate summaries or structured SOAP notes from a conversation between a doctor and a patient. We tackle this problem by generating section-wise summaries, classifying these summaries into appropriate section headers and generating full-length summaries from longer conversations.

We conclude from the results that prompting-based techniques by themselves can not perform optimally themselves but do give some outstanding results when combined with existing techniques, like prompt selection using MiniLM. Next, we also dive deep into where prompting-based methods underperform the standard models like BART and T5. 

Overall, our model concluded 3rd amongst all runs submitted and 2nd as a team for task A, which aimed at producing section-level summaries. Our system also stood 4th amongst all runs submitted and 2nd as a team in division-wise summaries for task B. In the future, we plan to use an ensemble of extractive and abstractive methods of generating summaries as well as using diversity algorithms that will aid in producing SOAP notes that are more robust and apt as per human evaluations.

\section{Limitations}
Considering the critical nature of the domain of the task, it is of paramount importance to ensure stability in the results expected from the model. Despite setting the temperature (T) as 0 for all decodings in our experiments, we observe the variance in the generated summaries across runs.  Table \ref{tab:res_stability} contains the results for three runs for Task A and Task B. The in-context examples for each sample and other parameters have been kept constant across these runs to identify the degree of stochasticity. Further, in-context learning has shown to be susceptible to changes in order of in-context examples \cite{lu2021fantastically}, as well as the template of the examples \cite{shin2020autoprompt}. A more reliable process to generate the summaries along with identification of the optimal examples (template, order) is thus required. Additionally, due to the context limit of the GPT-4 model, evaluating the impact of natural language instructions in addition to the examples could not be performed. 

\section{Ethics Statement}
There exist several risks and ethical considerations that necessitate comprehensive addressal prior to the deployment and utilization of our proposed methods utilizing Language Models (LLMs). A significant apprehension associated with employing LLMs for summarization, as evidenced during experimentation, is their susceptibility to hallucination.  This means that there would need to be stringent and effective fact-verification post-processing on the generated summaries, thereby ensuring their factual accuracy and alignment with the doctor-patient discourse.
\par The preservation of patient confidentiality and privacy assumes paramount importance within the context of healthcare data, given its highly sensitive and personal nature. Consequently, it becomes imperative to undertake effective data anonymization techniques to safeguard patient identities. Additionally, obtaining explicit consent from patients regarding the utilization of their data assumes critical significance. In tandem, strict adherence to the standards set forth by the Health Insurance Portability and Accountability Act (HIPAA) is essential to ensure compliance and guarantee the privacy and security of patient information. \par Furthermore, another vital aspect that demands careful consideration is the explainability and interpretability when utilizing Language Models (LLMs) for medical summarization. It becomes essential to address the challenge of  comprehending the decision-making processes underlying their outputs. Particularly within the medical domain, where critical decisions are made based on these outputs, explainability is of great importance.
\bibliography{anthology,custom}
\bibliographystyle{acl_natbib}
\clearpage

\newpage
\appendix
\onecolumn

\section{Appendix}
\label{sec:appendix}
This appendix presents two tables - Table \ref{tab:cat} contains the categories and subcategories in which the dialogue is divided to create a SOAP note. Table \ref{tab:prompts} presents the prompts used by approaches for tasks A and B.

\begin{table}[h]
\caption{Categorization Scheme}
\label{tab:cat}
\centering
\begin{tabular}{c}
\bottomrule
HISTORY OF PRESENT ILLNESS                    \\ \toprule
Fam/Sochx {[}Family History/Social History{]} \\
Genhx {[}History of Present Illness{]}        \\
Pastmedicalhx {[}Past Medical History{]}      \\
CC {[}Chief Complain{]}                      \\
Pastsurgical {[}Past Surgical History{]}      \\
Allergy                                       \\
Gynhx {[}Gynecologic History{]}               \\
Other\_history                                \\
Immunizations                                 \\
Medications                                   \\ \bottomrule
PHYSICAL EXAM                                 \\ \toprule
ROS {[}Review of Systems{]}                   \\
Exam                                          \\ \bottomrule
RESULTS                                       \\ \toprule
Imaging                                       \\
Procedures                                    \\
Labs                                          \\ \bottomrule
ASSESSMENT AND PLAN                           \\ \toprule
Assessment                                    \\ 
Diagnosis                                     \\ 
Plan                                          \\ 
Edcourse {[}Emergency Department Course{]}    \\ 
Disposition                                   \\ 
Medications                                   \\ \toprule
\end{tabular}%
\end{table}

\begin{table*}[]
\centering
\caption{Prompt Templates}
\label{tab:prompts}
\resizebox{\textwidth}{!}{%
\begin{tabular}{|l|l|l|}
\hline
Prompting Approach &
  Model &
  Prompt(Example) \\ \hline
Zero-Shot &
  text-davinci-003 &
  \begin{tabular}[c]{@{}l@{}}"Summarize the following into a medical report having the following sections: \\ 'HISTORY OF PRESENT ILLNESS', 'PHYSICAL EXAM', 'RESULTS', 'ASSESSMENT AND PLAN'.\end{tabular} \\ \hline
\begin{tabular}[c]{@{}l@{}}Few-shot prompting \\ for section-wise summary\end{tabular} &
  text-davinci-003 &
  \begin{tabular}[c]{@{}l@{}}Prompt for PHYSICAL EXAM section($k$=5)\\ """\\ Dialogue: \\ Doctor: Breath in breath out, let me tap it and see. Well, your lungs sound clear. \\ Patient: Okay.\\ \\ \\ Summary: \\ CHEST: Lungs bilaterally clear to auscultation and percussion.\\ …..\\ …..\\ …..\\ Dialogue : \\ Doctor: Do you have any chest pain?\\ Patient: No, I don't.\\ Doctor: Any breathlessness?\\ Patient: Yes, I do get breathless only when I have to do some form of exertion \\ like walking a long time or running.\\ Doctor: Okay. How about any bowel issues?\\ Patient: No, I don't have any stomach problems except I have to go frequently to use the bathroom.\\ Doctor: Okay frequency. How about any prolonged bleeding issues or anything like that sort?\\ Patient: No nothing like that.\\ \\ \\ Summary : \\ He denies any chest pain.  He admits to exertional shortness of breath.  \\ He denies any GI problems as noted.  Has frequent urination as noted.  \\ He denies any bleeding disorders or bleeding history.\\ \\ Dialogue : \\ \{dialogue\}\\ \\ \\ Summary :\\ """\end{tabular} \\ \hline
Perspective Shift &
  \begin{tabular}[c]{@{}l@{}}text-davinci-003\\ gpt3.5-turbo\end{tabular} &
  \begin{tabular}[c]{@{}l@{}}2 staged prompting (perspective shift with turbo and summarization with davinci)\\ \\ PERSPECTIVE SHIFT = """\\ Convert the following into third person.\\ \\ \{dialogue\} \textbackslash{}\textbackslash{}\textbackslash\\ """\\ \\ PROMPT = """\\ Summarize the following into a medical report having the following sections: \\ "HISTORY OF PRESENT ILLNESS", "PHYSICAL EXAM", "RESULTS", \\ "ASSESSMENT AND PLAN" where each section is at least 60 words.\\ \\ \{third-person-perspective\}\textbackslash{}\textbackslash{}\textbackslash\\ """\end{tabular} \\ \hline
Two-Stage Prompting &
  text-davinci-003 &
  \begin{tabular}[c]{@{}l@{}}PROMPT \#1\\ """\\ \{dialogue\}\\ \textbackslash{}\textbackslash\\ \\ List the important points from the above conversation for a medical report\\ """\\ \\ PROMPT \#2\\ """\\ \{prompt1-generated-output\}\\ \textbackslash{}\textbackslash\\ Create a paragraph from the above facts only\\ """\end{tabular} \\ \hline
\end{tabular}%
}
\end{table*}

\begin{table*}[]
\centering
\resizebox{\textwidth}{!}{%
\begin{tabular}{|l|l|l|}
\hline
Prompting Approach &
  Model &
  Prompt(Example) \\ \hline
\begin{tabular}[c]{@{}l@{}}Prompt Selection - \\ MMR($k$=3)\end{tabular} &
  text-davinci-003 &
  \begin{tabular}[c]{@{}l@{}}PROMPT SELECTION with $k$=3\\ Dialogue:\\ Doctor: Your last visit was on April seventh two thousand five, correct? \\ Patient: Ah no, it was on April eighth two thousand five, doctor. \\ Doctor: That's right. So, has anything changed since then?\\ Patient: No, everything is the same really.\\ \\ Summary:\\ Essentially unchanged from my visit of 04/08/2005.\\ …..\\ …..\\ …..\\ Dialogue:\\ Doctor: Do you have any past or present medical conditions? \\ Patient: No.\\ \\ Summary:\\ None.\\ …..\\ …..\\ …..\\ Dialogue:\\ \{dialogue\}\\ \\ Summary:\end{tabular} \\ \hline
\textbf{\begin{tabular}[c]{@{}l@{}}Prompt Selection - \\ Semantic Similarity($k$=7) - Task A\end{tabular}} &
  \textbf{GPT-4} &
  \begin{tabular}[c]{@{}l@{}}PROMPT SELECTION with $k$=7\\ Dialogue:\\ Doctor: Do you know about any medical issues running in your family?\\ Patient: Yeah, almost everyone had diabetes.\\ \\ Summary:\\ Multiple family members have diabetes mellitus.\\ …..\\ …..\\ …..\\ Dialogue:\\ Doctor: Any specific family medical history that I should be aware of? \\ Patient: No. \\ Doctor: Anyone in your family, even grandparents, if you know them, did\\ they have diabetes or high blood pressure? \\ Patient: No. \\ Doctor: Anyone else sick at home? \\ Patient: No.\\ \\ Summary:\\ Noncontributory.  No one else at home is sick.\\ …..\\ …..\\ …..\\ Dialogue:\\ \{dialogue\}\\ \\ Summary:\end{tabular} \\ \hline
\textbf{\begin{tabular}[c]{@{}l@{}}Prompt Selection - \\ Semantic Similarity($k$=1) Task B\end{tabular}} &
  \textbf{GPT-4} &
  \begin{tabular}[c]{@{}l@{}}PROMPT SELECTION with $k$=1\\ Dialogue:\\ {[}doctor{]} and why is she here ? annual exam. okay. all right. hi, Sarah. how are you ?\\ {[}patient{]} good . how are you ?\\ {[}doctor{]} i'm good . are you ready to get started ?\\ {[}patient{]} yes , i am .\\ {[}doctor{]} okay . so Sarah is a 27-year-old female here for her annual visit. \\ So, Sarah, how have you been since the last time I saw you ?\\ …….\\ …….\\ Summary:\\ CHIEF COMPLAINT\\ Annual visit.\\ HISTORY OF PRESENT ILLNESS\\ The patient is a 27-year-old female who presents for her annual visit. \\ She reports that she has been struggling with her depression off and on for the past year...... \\ …….\\ \\ Dialogue:\\ \{dialogue\}\\ \\ Summary: \\ \end{tabular} \\ \hline
\end{tabular}%
}
\end{table*}

\end{document}